\documentclass[runningheads]{llncs}
\usepackage{graphicx}
\usepackage{amsmath,amssymb} 
\usepackage{color}

\usepackage{bm}
\usepackage{multirow}
\usepackage{hyperref}

\graphicspath{{figure/}}

\begin{document}

\newcommand{\point}{
    \raise0.7ex\hbox{.}
    }


\pagestyle{headings}

\mainmatter

\title{Learning a Mixture of Deep Networks for Single Image Super-Resolution} 

\titlerunning{Learning a Mixture of Deep Networks for Single Image Super-Resolution} 

\authorrunning{Ding Liu, Zhaowen Wang, Nasser Nasrabadi, and Thomas Huang} 

\author{Ding Liu\textsuperscript{\dag}, Zhaowen Wang\textsuperscript{\ddag},  Nasser Nasrabadi\textsuperscript{\S}, and Thomas Huang\textsuperscript{\dag}} 
\institute{\textsuperscript{\dag}Beckman Institute,  University of Illinois at Urbana-Champaign, IL, USA \\
	\textsuperscript{\ddag}Adobe Research, CA, USA \\
	\textsuperscript{\S}Lane Department of Computer Science and Electrical Engineering,  West Virginia University, WV, USA
} 
\maketitle

\begin{abstract}
Single image super-resolution (SR) is an ill-posed problem which aims to recover high-resolution (HR) images from their low-resolution (LR) observations. The crux of this problem lies in learning the complex mapping between low-resolution patches and the corresponding high-resolution patches. 
Prior arts have used either a mixture of simple regression models or a single non-linear neural network for this propose. 
This paper proposes the method of learning a mixture of SR inference modules in a unified framework to tackle this problem. 
Specifically, a number of SR inference modules specialized in different image local patterns are first independently applied on the LR image to obtain various HR estimates, 
and the resultant HR estimates are adaptively aggregated to form the final HR image. By selecting neural networks as the SR inference module, the whole procedure can be incorporated into a unified network and be optimized jointly.
Extensive experiments are conducted to investigate the relation between restoration performance and different network architectures. Compared with other current image SR approaches, our proposed method achieves state-of-the-arts restoration results on a wide range of images consistently while allowing more flexible design choices. The source codes are available in \url{http://www.ifp.illinois.edu/~dingliu2/accv2016}.
\end{abstract}


\section{Introduction}

\label{sec:intro}

Single image super-resolution (SR) is usually cast as an inverse problem of recovering the original high-resolution (HR) image from the low-resolution (LR) observation image. This technique can be utilized in the applications where high resolution is of importance, such as photo enhancement, satellite imaging and SDTV to HDTV conversion \cite{park2003super}. The main difficulty resides in the loss of much information in the degradation process. Since the known variables from the LR image is usually greatly outnumbered by that from the HR image, this problem is a highly ill-posed problem. 

A large number of single image SR methods have been proposed in the literature,
including interpolation based method \cite{morse2001image}, edge model based method \cite{fattal2007image} and example based method \cite{chang2004super,glasner2009super,yang2010image,timofte2013anchored,dong2015image,huang2015single}. 
Since the former two methods usually suffer the sharp drop in restoration performance with large upscaling factors, the example based method draws great attention from the community recently. It usually learns the mapping from LR images to HR images in a patch-by-patch manner, with the help of sparse representation \cite{yang2010image,wang2015learning}, random forest \cite{schulter2015fast} and so on. 
The neighbor embedding method \cite{chang2004super,timofte2013anchored} and neural network based method \cite{dong2015image} are two representatives of this category.

Neighbor embedding is proposed in \cite{chang2004super,bevilacqua2012low} which estimates HR patches as a weighted average of local neighbors with the same weights as in LR feature space,
based on the assumption that LR/HR patch pairs share similar local geometry in low-dimensional nonlinear manifolds. 
The coding coefficients are first acquired by representing each LR patch as a weighted average of local neighbors, and then the HR counterpart is estimated by the multiplication of the coding coefficients with the corresponding training HR patches. 
Anchored neighborhood regression (ANR) is utilized in \cite{timofte2013anchored} to improve the neighbor embedding methods, which partitions the feature space into a number of clusters using the learned dictionary atoms as a set of anchor points. 
A regressor is then learned for each cluster of patches. This approach has demonstrated superiority over the counterpart of global regression in \cite{timofte2013anchored}. 
Other variants of learning a mixture of SR regressors can be found in \cite{timofte2014a+,dai2015jointly,timofte2016seven}.



Recently, neural network based models have demonstrated the strong capability  for single image SR \cite{cui2014deep,dong2015image,wang2015deep}, due to its large model capacity and the end-to-end learning strategy to get rid of hand-crafted features.
Cui et al. \cite{cui2014deep} propose using a cascade of stacked collaborative local autoencoders for robust matching of self-similar patches, in order to increase the resolution of inputs gradually. Dong et al. \cite{dong2015image} exploit a fully convolutional neural network (CNN) to approximate the complex non-linear mapping between the LR image and the HR counterpart.
A neural network that closely mimics the sparse coding approach for image SR is designed by Wang et al \cite{wang2015deep,liu2016robust}.
Kim et al proposes a very deep neural network with residual architecture to exploit contextual information over large image regions \cite{kim2016accurate}.


In this paper, we propose a method to combine the merits of the neighborhood embedding methods and the neural network based methods via learning a mixture of neural networks for single image SR.
The entire image signal space can be partitioned into several subspaces, and we dedicate one SR module  to the image signals in each subspace, the synergy of which allows a better capture of the complex relation between the LR image signal and its HR counterpart  than the generic model.
In order to take advantage of the end-to-end learning strategy of neural network based methods, we choose neural networks as the SR inference modules and incorporate these modules into one unified network, and design a branch in the network to predict the pixel-level weights for HR estimates from each SR module before they are adaptively aggregated to form the final HR image.

A systematic analysis of different network architectures is conducted with the focus on the relation between SR performance and various network architectures via extensive experiments, where the benefit of utilizing a mixture of SR models is demonstrated. Our proposed approach is contrasted with other current popular approaches on a large number of test images, and achieves state-of-the-arts performance consistently along with more flexibility of model design choices.

The paper is organized as follows. The proposed method is introduced and explained in detail in Section \ref{sec:proposed}.  Section \ref{sec:exp} describes our experiments, in which we analyze thoroughly different network architectures and compare the performance of our method with other current SR methods both quantitatively and qualitatively. Finally in Section \ref{sec:conclusions} we conclude the paper. 

\section{Proposed Method}

\label{sec:proposed}

\begin{figure*}[t]
	\center
	\includegraphics[width=0.8\linewidth]{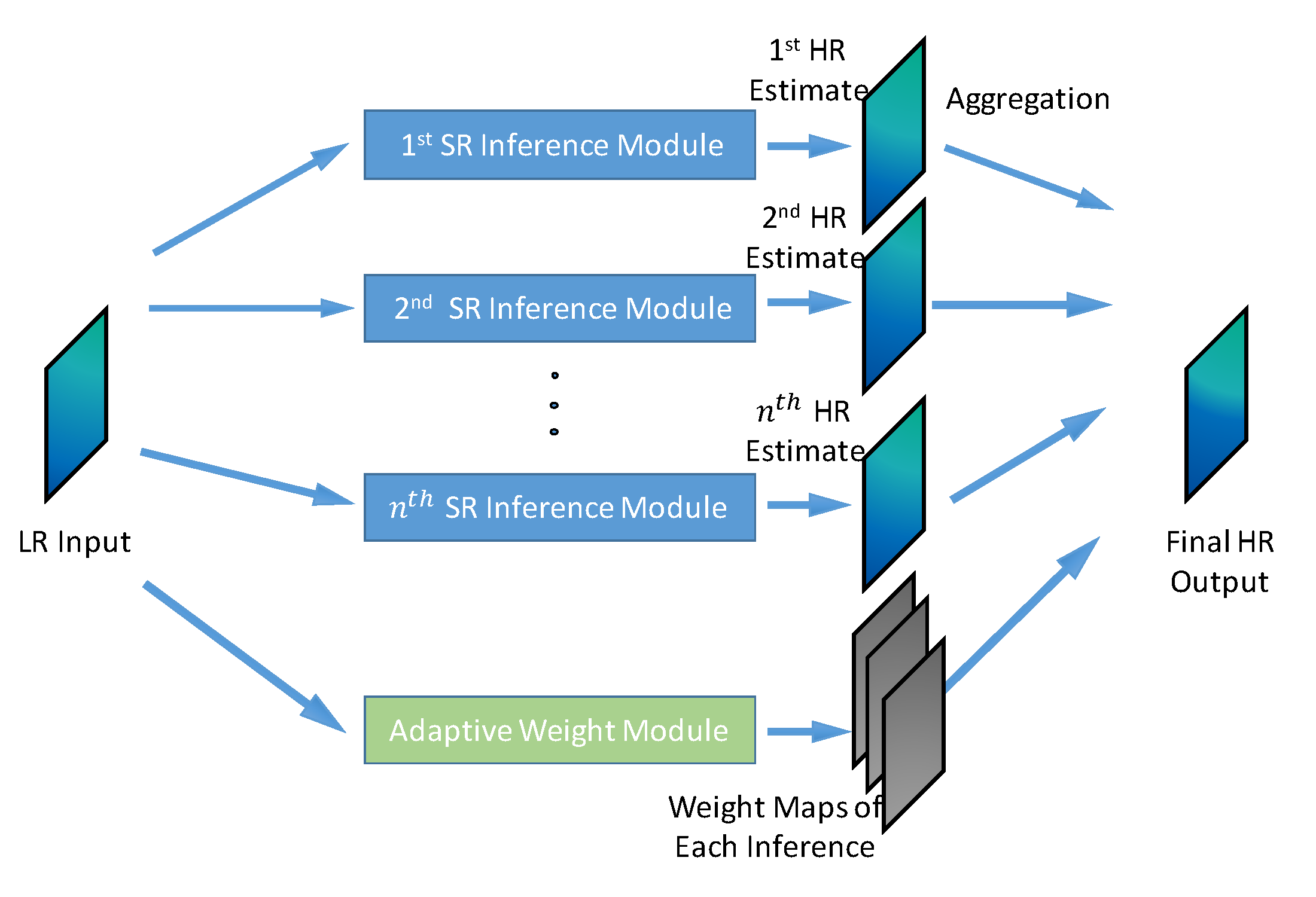}
	\caption{The overview of our proposed method. It consists of a number of SR inference modules and an  adaptive weight module. Each SR inference module  is dedicated to inferencing a certain class of image local patterns, and is independently applied on the LR image to predict one HR estimate. These estimates are adaptively combined using pixel-wise aggregation weights from the adaptive weight module in order to form the final HR image.}
	\label{fig:overview}
\end{figure*}

\subsection{Overview}

First we give the overview of our method. The LR image serves as the input to our method.
There are a number of \textbf{SR inference modules} $\{B_i\}_{i=1}^N$ in our method. Each of them, $B_i$, is dedicated to inferencing a certain class of image patches, and applied on the LR input image to predict a HR estimate. 
We also devise an \textbf{adaptive weight module}, $T$, to adaptively combine at the pixel-level the HR estimates from SR inference modules.
When we select neural networks as the SR inference modules, all the components can be incorporated into a unified neural network and be jointly learned.
The final estimated HR image is adaptively aggregated from the estimates of all SR inference modules.
The overview of our method is shown in Figure \ref{fig:overview}.

\subsection{Network Architecture}

\subsubsection{SR Inference Module}
\begin{figure}
	\center
	\includegraphics[width=0.8\linewidth]{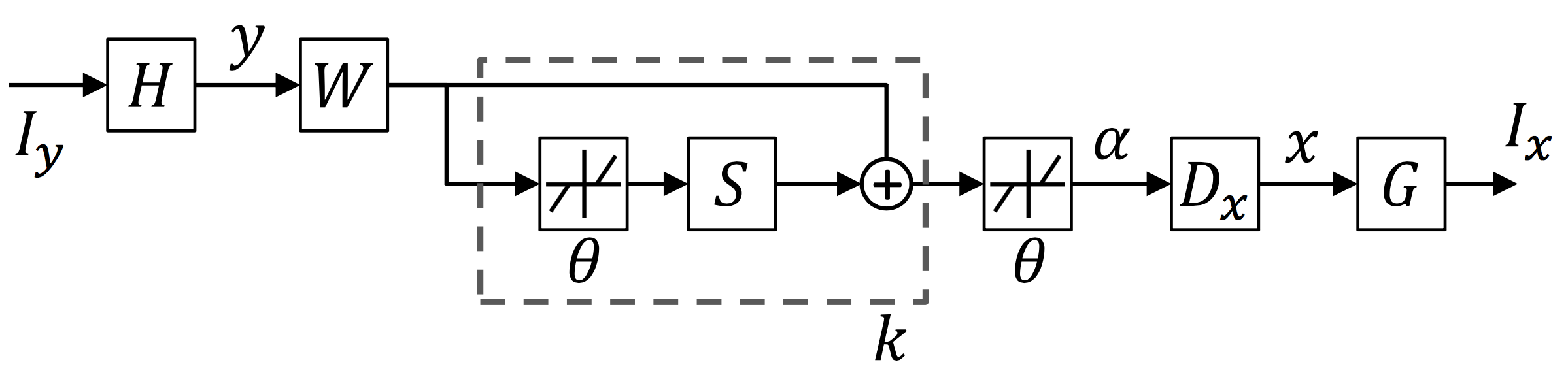}
	\caption{The network architecture of SCN \cite{wang2015deep}, which serves as the SR inference module in our method.}
	\label{fig:scn}
\end{figure}

: 
Taking the LR image as input, each SR inference module is designed to better capture the complex relation between a certain class of LR image signals and its HR counterpart, while predicting a HR estimate.
For the sake of inference accuracy, we choose as the SR inference module a recent sparse coding based network (SCN) in \cite{wang2015deep}, 
which implicitly incorporates the sparse prior into neural networks via employing the learned iterative shrinkage and thresholding algorithm (LISTA), and closely mimics the sparse coding based image SR method \cite{yang2012coupled}.
The architecture of SCN is shown in Figure \ref{fig:scn}. 
Note that the design of the SR inference module is not limited to SCN, and all other neural network based SR models, e.g. SRCNN \cite{dong2015image}, can work as the SR inference module as well.
The output of $B_i$ serves as an estimate to the final HR frame.

\subsubsection{Adaptive Weight Module}:
The goal of this module is to model the selectivity of  the HR estimates from every SR inference module.
We propose assigning pixel-wise aggregation weights of each HR estimate,  
and again the design of this module is open to any operation in the field of neural networks.
Taking into account the computation cost and efficiency, we utilize only three convolutional layers for this module, and 
ReLU is applied on the filter responses to introduce non-linearity.
This module finally outputs the pixel-level weight maps for all the HR estimates.

\subsubsection{Aggregation}: 
Each SR inference module's output is pixel-wisely multiplied with its corresponding weight map from the adaptive weight module, and then these products are summed up to form the final estimated HR frame.
If we use $\mathbf{y}$ to denote the LR input image, a function $W(\mathbf{y}; \theta_w)$ with parameters $\theta_w$ to represent the behavior of the adaptive weight module, and a function $F_{B_i}(\mathbf{y}; \theta_{B_i})$ with parameters $\theta_{B_i}$ to represent the output of SR inference module $B_i$, the final estimated HR image $F(\mathbf{y}; \mathbf{\Theta})$ can be expressed as

\begin{equation}
F(\mathbf{y}; \mathbf{\Theta}) = \sum_{i=1}^{N} W_i(\mathbf{y}; \theta_w) \odot F_{B_i}(\mathbf{y}; \theta_{B_i}), 
\label{eqn:weightmodule}
\end{equation}

\noindent
where $\odot$ denotes the point-wise multiplication.

\subsection{Training Objective}

In training, our model tries to minimize the loss between the target HR frame and the predicted output, as

\begin{equation}
％\vspace{-2mm}
\min\limits_{\mathbf{\Theta}} \sum_{j} \|F(\mathbf{y}_j; \mathbf{\Theta}) - \mathbf{x}_j \|^2_2,
\label{eqn:cost}
\end{equation} 
\noindent
where $F(\mathbf{y}; \mathbf{\Theta})$ represents the output of our model, $\mathbf{x}_j$ is the $j$-th HR image and $\mathbf{y}_j$ is the corresponding LR image;
$\mathbf{\Theta}$ is the set of all parameters in our model.

If we plug Eqn. \ref{eqn:weightmodule} into Eqn. \ref{eqn:cost}, 
the cost function then can be expanded as:

\begin{equation}
\min\limits_{ \theta_w, \{\theta_{B_i}\}_{i=1}^{N}} \sum_{j} \| \sum_{i=1}^{N} W_i(\mathbf{y}_j; \theta_w) \odot F_{B_i}(\mathbf{y}_j; \theta_{B_i}) - \mathbf{x}_j \|^2_2.
\label{eqn:formulation}
\end{equation}


\section{Experiments}

\label{sec:exp}

\subsection{Data Sets and Implementation Details}

We conduct experiments following the protocols in \cite{timofte2013anchored}. Different learning based methods use different training data in the literature. We choose 91 images proposed in \cite{yang2010image} to be consistent with \cite{timofte2014a+,schulter2015fast,wang2015deep}. These training data are augmented with translation, rotation and scaling, providing approximately 8 million training samples of $56 \times 56$ pixels. 

Our model is tested on three benchmark data sets, which are Set5 \cite{bevilacqua2012low}, Set14 \cite{zeyde2012single} and BSD100 \cite{martin2001database}. The ground truth images are downscaled by bicubic interpolation to generate LR/HR image pairs for both training and testing.

Following the convention in \cite{timofte2013anchored,wang2015deep}, we convert each color image into the YCbCr colorspace and only process the luminance channel with our model, and bicubic interpolation is applied to the chrominance channels, because the visual system of human is more sensitive to details in intensity than in color.

Each SR inference module adopts the network architecture of SCN, while the filters of all three convolutional layers in the adaptive weight module have the spatial size of $5 \times 5$ and the numbers of filters of three layers are set to be $32, 16$ and $N$, which is the number of SR inference modules.

Our network is implemented using Caffe \cite{jia2014caffe} and is trained on a machine with 12 Intel Xeon 2.67GHz CPUs and 1 Nvidia TITAN X GPU. For the adaptive weight module, we employ a constant learning rate of $10^{-5}$ and initialize the weights from Gaussian distribution, while we stick to the learning rate and the initialization method in \cite{wang2015deep} for the SR inference modules. The standard gradient descent algorithm is employed to train our network with a batch size of 64 and the momentum of 0.9. 

We train our model for the upscaling factor of 2. For larger upscaling factors,  we adopt the model cascade technique in \cite{wang2015deep} to apply $\times 2$ models multiple times until the resulting image reaches at least as large as the desired size. The resulting image is downsized via bicubic interpolation to the target resolution if necessary.

\begin{table}
	\centering
	\caption{PSNR (in dB) and SSIM comparisons on Set5, Set14 and BSD100 for $\times 2$, $\times 3$ and $\times 4$ upscaling factors among various network architectures.
		\textcolor{red}{Red} indicates the best and \textcolor{blue}{blue} indicates the second best performance.
	}
	\label{tab:psnr}
	\begin{tabular}{|c|c||c|c|c|}
		\hline
		\multicolumn{2}{|c||}{\multirow{2}{*}{Benchmark}} &  SCN  & MSCN-2  & MSCN-4   \\
		\multicolumn{2}{|c||}{ } & ($n=128$)  & ($n=64$) & ($n=32$)  \\
		\hline
		\hline
		\multirow{3}{*}{Set5} & $\times 2$ &  36.93 / 0.9552 & \textcolor{red}{37.00} / \textcolor{blue}{0.9558} &  \textcolor{blue}{36.99} / \textcolor{red}{0.9559} \\
		& $\times 3$ &  33.10 / \textcolor{red}{0.9136} & \textcolor{red}{33.15} / \textcolor{blue}{0.9133} & \textcolor{blue}{33.13} / 0.9130 \\
		& $\times 4$ &  30.86 / \textcolor{blue}{0.8710} & \textcolor{blue}{30.92} / 0.8709 & \textcolor{red}{30.93} / \textcolor{red}{0.8712} \\
		\hline
		\multirow{3}{*}{Set14} & $\times 2$ & 32.56 / 0.9069 & \textcolor{blue}{32.70} / \textcolor{blue}{0.9074} & \textcolor{red}{32.72} / \textcolor{red}{0.9076} \\
		& $\times 3$ & 29.41 / 0.8235 & \textcolor{blue}{29.53} / \textcolor{blue}{0.8253} & \textcolor{red}{29.56} / \textcolor{red}{0.8256} \\
		& $\times 4$ & 27.64 / 0.7578 & \textcolor{blue}{27.76} / \textcolor{blue}{0.7601} &  \textcolor{red}{27.79} / \textcolor{red}{0.7607} \\		
		\hline
		\multirow{3}{*}{BSD100} & $\times 2$ & 31.40 / 0.8884 & \textcolor{blue}{31.54} / \textcolor{blue}{0.8913} & \textcolor{red}{31.56} / \textcolor{red}{0.8914} \\
		& $\times 3$ & 28.50 / 0.7885 & \textcolor{blue}{28.56} / \textcolor{blue}{0.7920} &  \textcolor{red}{28.59} / \textcolor{red}{0.7926}  \\
		& $\times 4$ & 27.03 / 0.7161 & \textcolor{blue}{27.10} / \textcolor{blue}{0.7207} &  \textcolor{red}{27.13} / \textcolor{red}{0.7216} \\		
		\hline
	\end{tabular}
\end{table}

\subsection{SR Performance vs. Network Architecture}

In this section we investigate the relation between various numbers of SR inference modules and SR performance. 
For the sake of our analysis,  we increase the number of inference modules as  we decrease the module capacity of each of them, so that the total model capacity is approximately consistent and thus the comparison is fair.
Since the  chosen SR inference module, SCN \cite{wang2015deep}, closely mimics the sparse coding based SR method, we can reduce the module capacity of each inference module by decreasing the embedded dictionary size $n$ (i.e. the number of filters in SCN), for sparse representation.
We compare the following cases: 
\begin{itemize}
	\item one inference module with $n=128$, which is equivalent to the structure of SCN in \cite{wang2015deep}, denoted as \textit{SCN (n=128)}. Note that there is no need to include the adaptive weight module in this case. 
	\item two inference modules with $n=64$, denoted as \textit{MSCN-2 (n=64)}.
	\item four inference modules with $n=32$, denoted as \textit{MSCN-4 (n=32)}.
\end{itemize}
The average Peak Signal-to-Noise Ratio (PSNR) and structural similarity (SSIM) \cite{wang2004image} are measured  to quantitatively compare the SR performance of these models over Set5, Set14 and BSD100 for various upscaling factors ($\times 2, \times 3, \times 4$), and the results are displayed in Table \ref{tab:psnr}.

It can be observed that \textit{MSCN-2 (n=64)} usually outperforms the original SCN network, i.e. \textit{SCN (n=128)}, and \textit{MSCN-4 (n=32)} can achieve the best SR performance by improving the performance marginally over \textit{MSCN-2 (n=64)}.  
This demonstrates the effectiveness of our approach 
that each SR inference model is able to super-resolve its own class of image signals better than one single generic inference model.

\begin{figure*}
	\center
	\begin{tabular}{p{2mm}@{\hskip 2mm}c@{\hskip 1mm}c@{\hskip 1mm}c}
		\rotatebox{90}{\hspace{3.5mm} Weight map 1} &
		\includegraphics[height=0.24\linewidth]{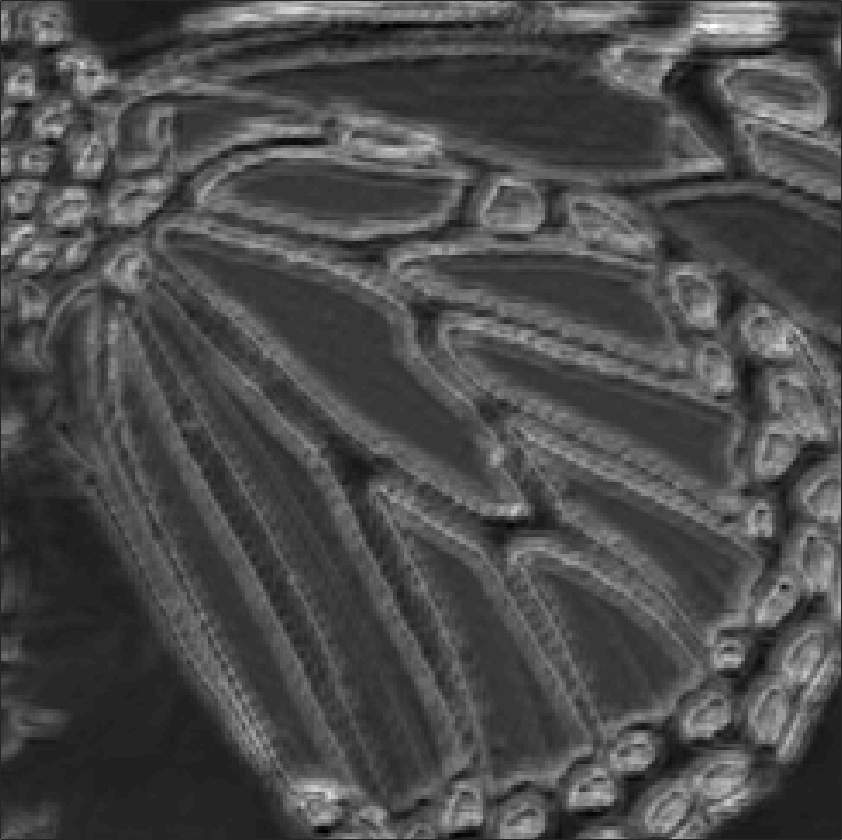} &
		\includegraphics[height=0.24\linewidth]{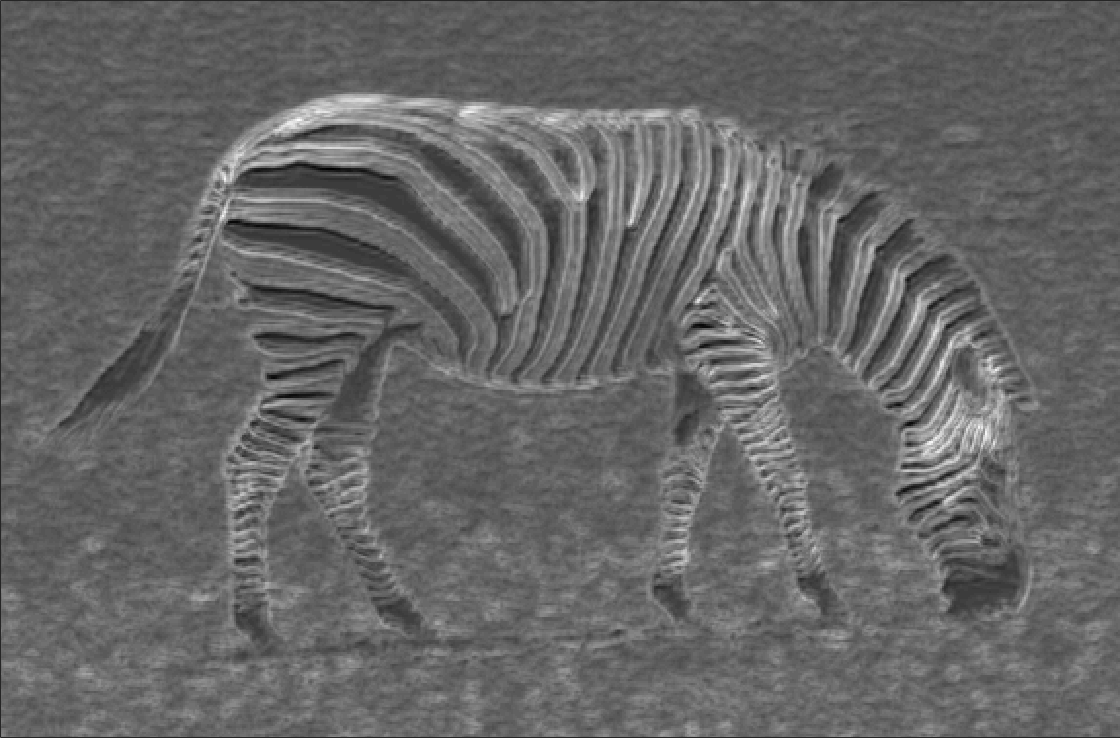} &
		\includegraphics[height=0.24\linewidth]{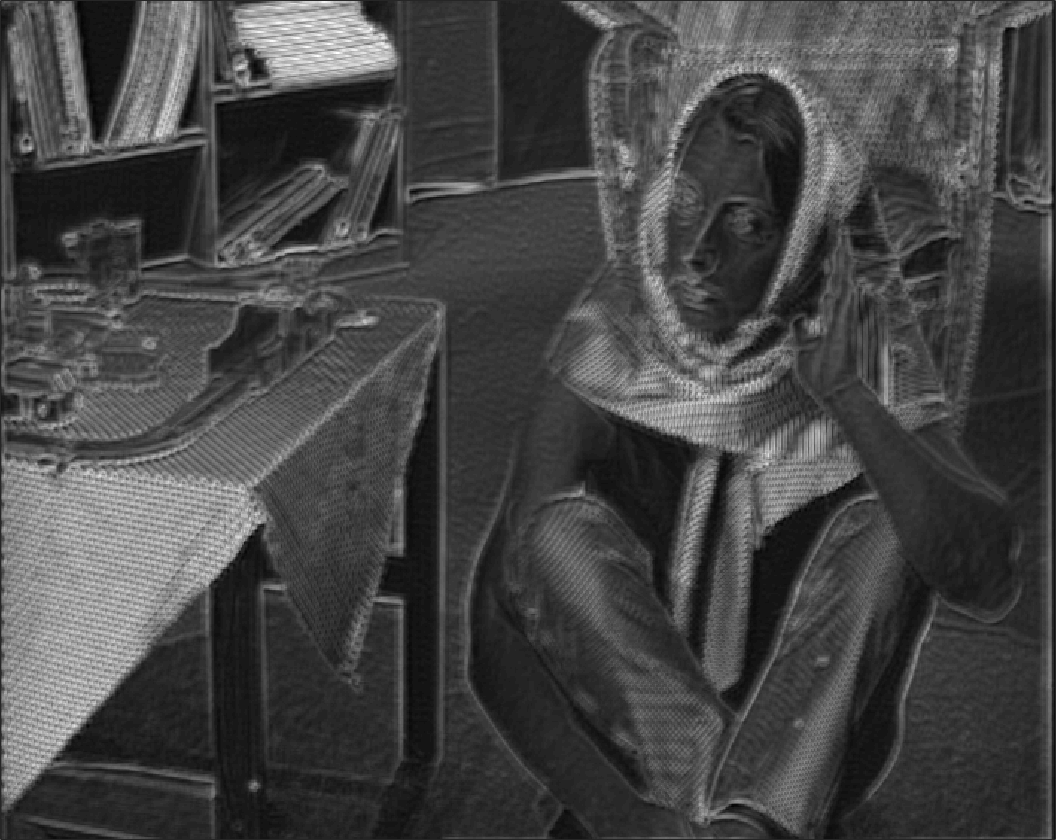} \\
		\rotatebox{90}{\hspace{3.5mm} Weight map 2} &
		\includegraphics[height=0.24\linewidth]{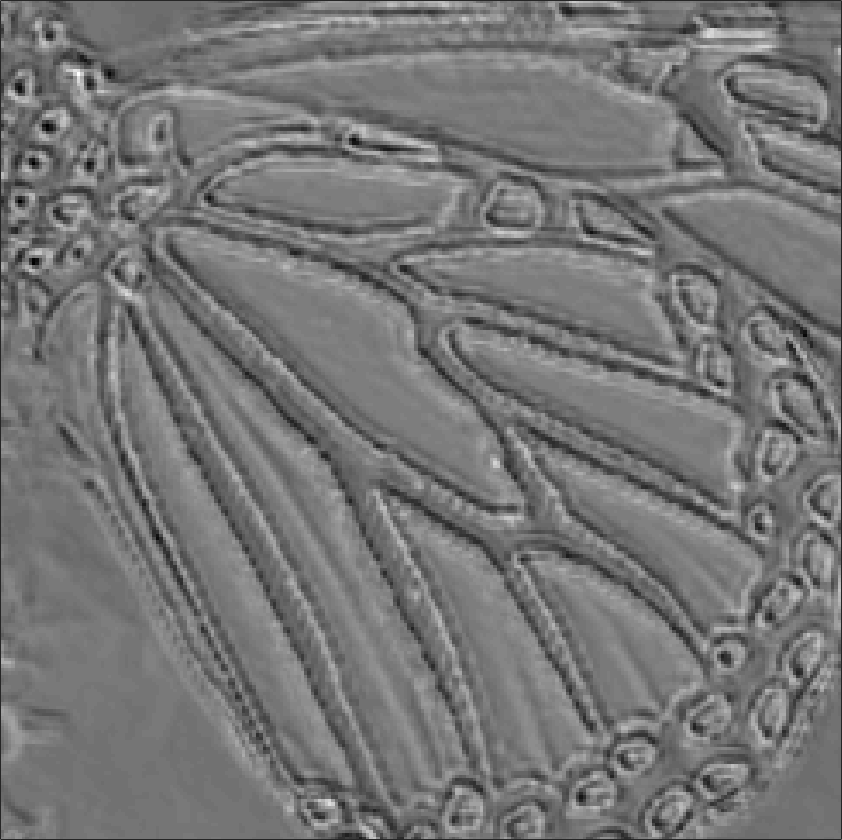} &
		\includegraphics[height=0.24\linewidth]{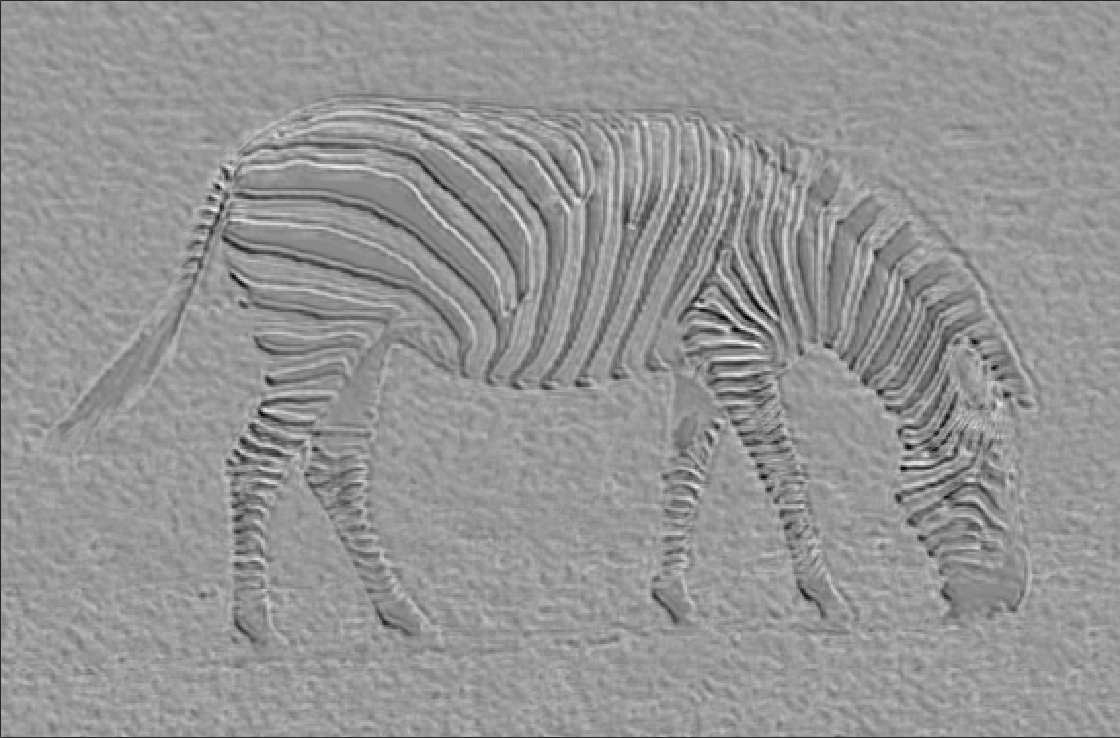} &
		\includegraphics[height=0.24\linewidth]{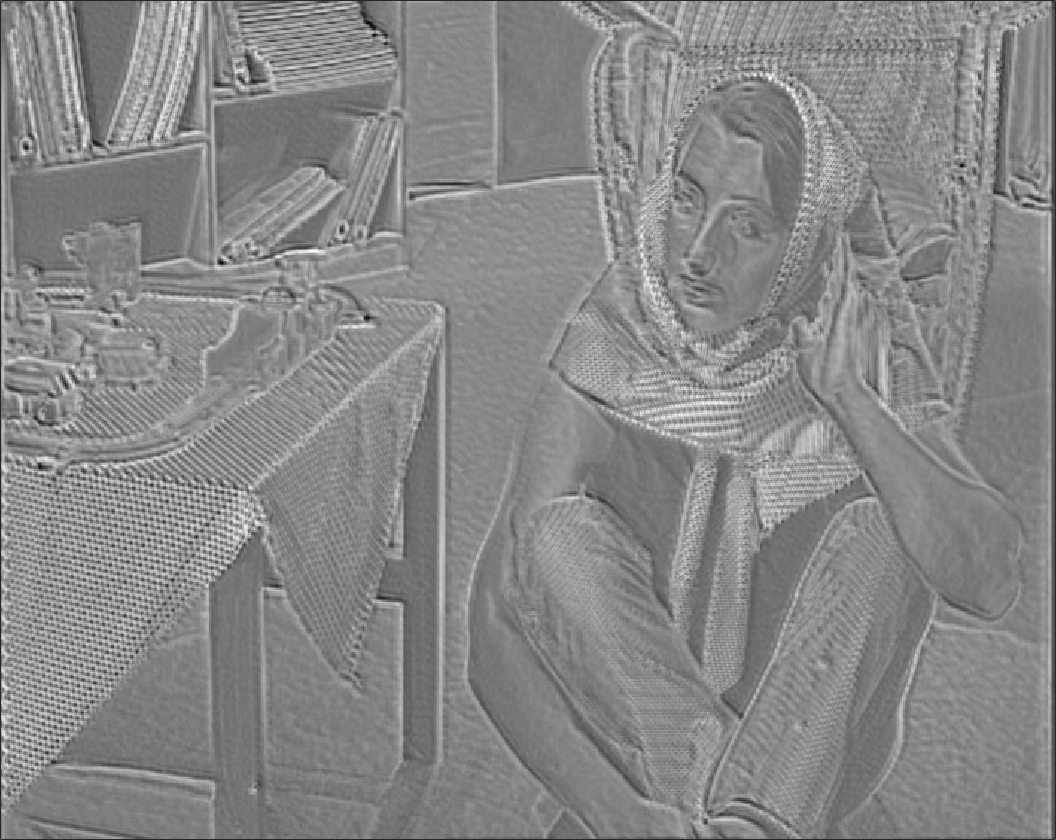} \\
		\rotatebox{90}{\hspace{3.5mm} Weight map 3} &
		\includegraphics[height=0.24\linewidth]{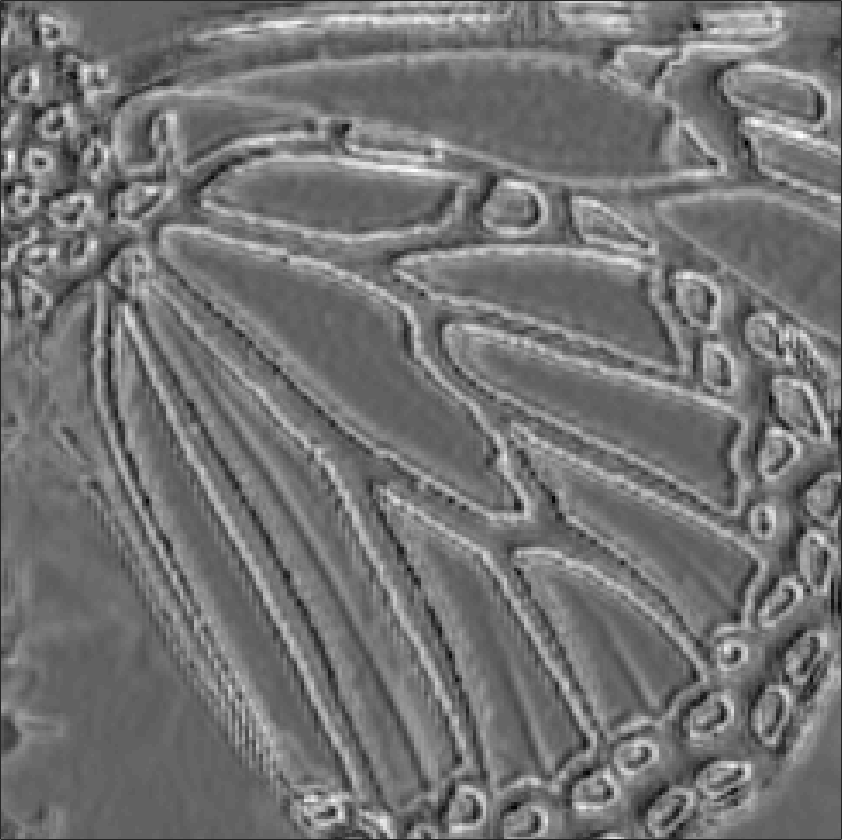} &
		\includegraphics[height=0.24\linewidth]{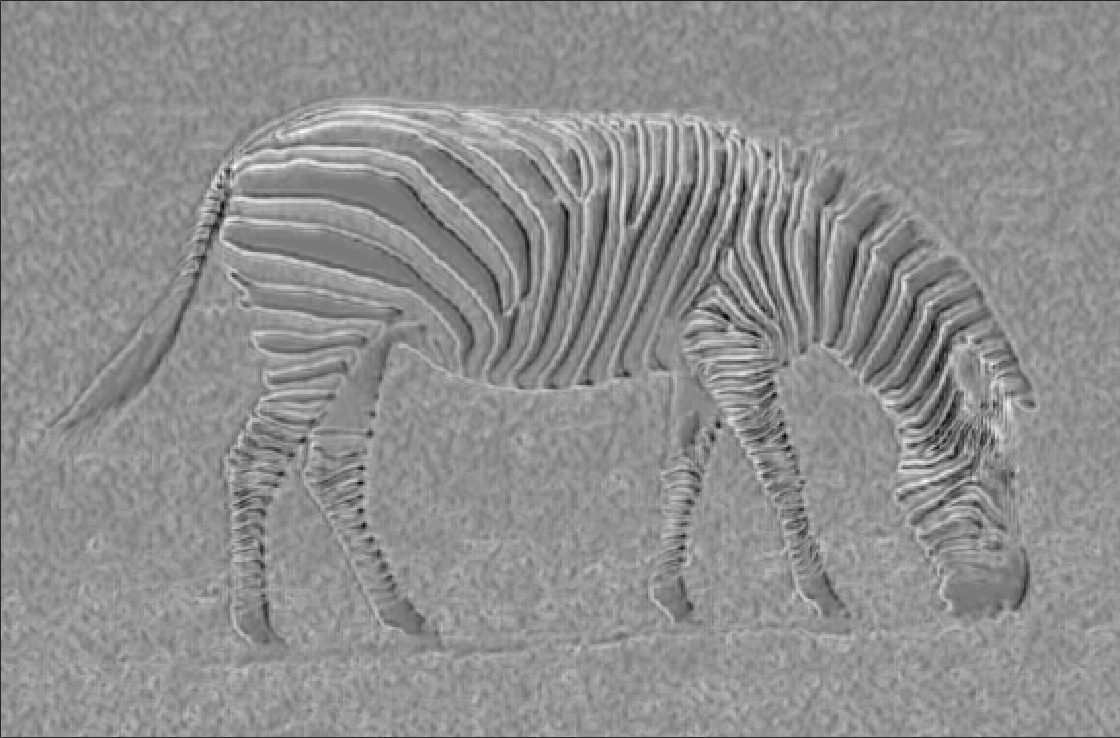} &
		\includegraphics[height=0.24\linewidth]{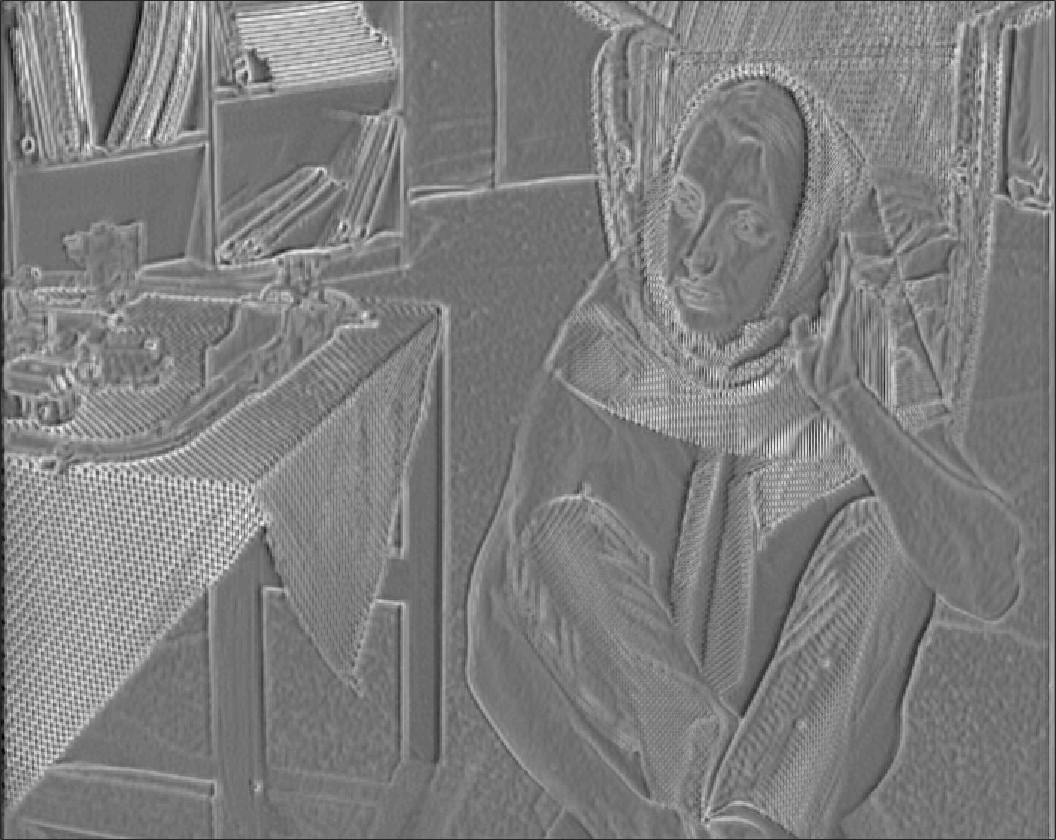} \\
		\rotatebox{90}{\hspace{3.5mm} Weight map 4} &
		\includegraphics[height=0.24\linewidth]{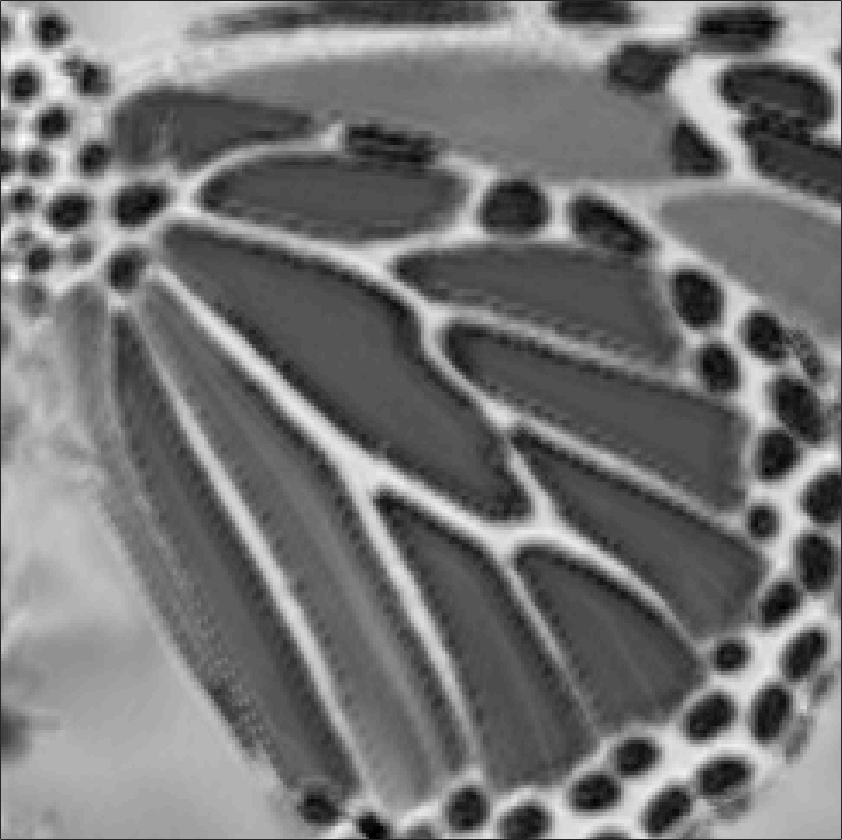} &
		\includegraphics[height=0.24\linewidth]{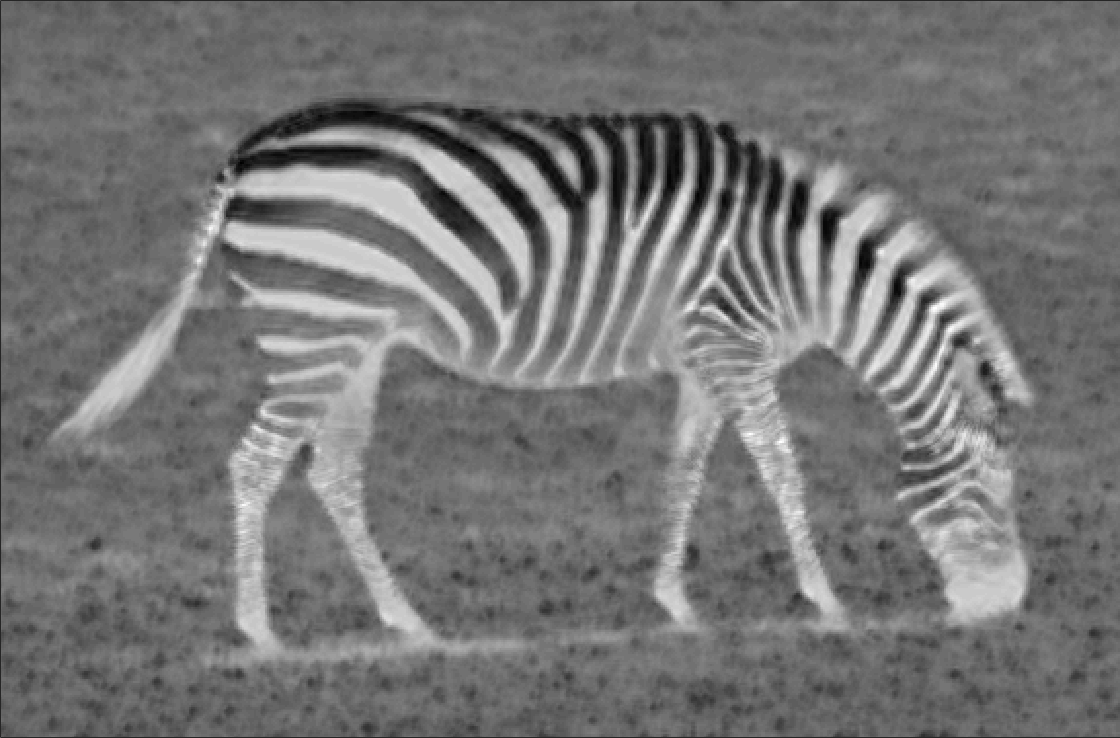} &
		\includegraphics[height=0.24\linewidth]{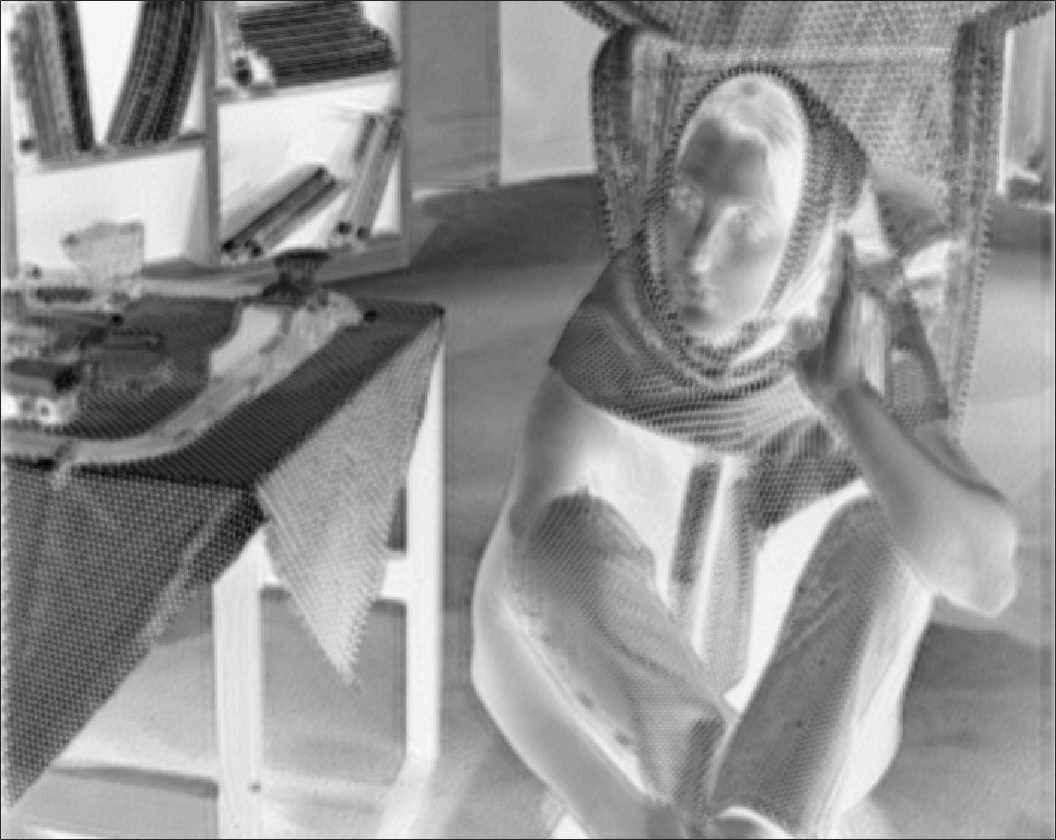} \\
		\rotatebox{90}{\hspace{3.5mm} Max label map} &
		\includegraphics[height=0.24\linewidth]{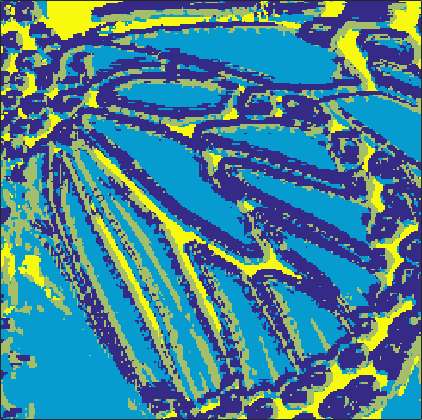} &
		\includegraphics[height=0.24\linewidth]{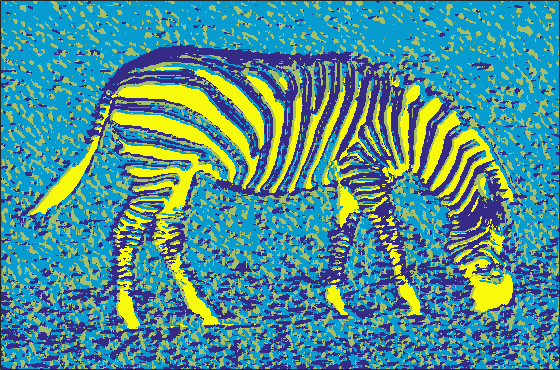} &
		\includegraphics[height=0.24\linewidth]{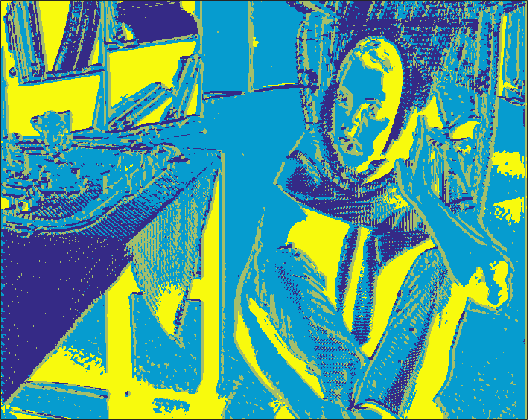} \\
	\end{tabular}
	\caption{ Weight maps for the HR estimate from every SR inference module in \textit{ MSCN-4} are given in the first four rows.
		The map (\textit{max label map}) which records the index of the maximum weight across all weight maps at every pixel is shown in the last row.
		Images from left to right: the \textit{butterfly} image upscaled by $\times 2$;
		the \textit{zebra} image upscaled by $\times2$; the \textit{barbara} image upscaled by $\times2$.}
	\label{fig:weightmap}
\end{figure*}

In order to further analyze the adaptive weight module, we select several input images, namely, \textit{butterfly}, \textit{zebra}, \textit{barbara},   and visualize the four weight maps for every SR inference module in the network. 
Moreover, we record the index of the maximum weight across all weight maps at every pixel and generate a \textit{max label map}.
These results are displayed in Figure \ref{fig:weightmap}.

From these visualizations it can be seen that weight map 4 shows high responses in many uniform regions, and thus mainly contributes to the low frequency regions of HR predictions. On the contrary, weight map 1, 2 and 3 have large responses in regions with various edges and textures, and restore the high frequency details of HR predictions. These weight maps reveal that these sub-networks work in a supplementary manner for constructing the final HR predictions. 
In the \textit{max label map}, similar structures and patterns of images usually share with the same label, indicating that  such similar  textures and patterns are favored to be super-resolved by the same inference model.

\begin{table*}[t]
	\centering
	\caption{PSNR (SSIM) comparison on three test data sets for various upscaling factors among different methods.
		The best performance is indicated in \textcolor{red}{red} and the second best performance is shown in \textcolor{blue}{blue}.
		The performance gain of our best model over all the other models' best is shown in the last row.}
	\label{tab:psnr2}
	\resizebox{\textwidth}{!}{
	\begin{tabular}{r||c|c|c||c|c|c||c|c|c}
		\hline
		Data Set      & \multicolumn{3}{c||}{Set5}	& \multicolumn{3}{c||}{Set14}	& \multicolumn{3}{c}{BSD100} \\
		\hline
		Upscaling &	$\times2$ &	$\times3$ & $\times4$ &
		$\times2$ &	$\times3$ & $\times4$ &
		$\times2$ &	$\times3$ & $\times4$ \\
		\hline
		\hline
		\multirow{2}{*}{A+ \cite{timofte2014a+}} &
		36.55 & 32.59 & 30.29 & 32.28 & 29.13 & 27.33 & 31.21 & 28.29 & 26.82 \\
		& (0.9544) & (0.9088) & (0.8603) & (0.9056) & (0.8188) & (0.7491) & (0.8863) & (0.7835) & (0.7087) \\
		\hline
		\multirow{2}{*}{SRCNN \cite{dong2015image}} &
		36.66 & 32.75 & 30.49 & 32.45 & 29.30 & 27.50 & 31.36 & 28.41 & 26.90 \\
		& (0.9542) & (0.9090) & (0.8628) & (0.9067) & (0.8215) & (0.7513) & (0.8879) & (0.7863) & (0.7103) \\
		\hline
		\multirow{2}{*}{RFL \cite{schulter2015fast}} &
		36.54 & 32.43 & 30.14 & 32.26 & 29.05 & 27.24 & 31.16 & 28.22 & 26.75 \\
		& (0.9537) & (0.9057) & (0.8548) & (0.9040) & (0.8164) & (0.7451) & (0.8840) & (0.7806) & (0.7054) \\
		\hline
		\multirow{2}{*}{SelfEx \cite{huang2015single}} &
		36.49 & 32.58 & 30.31 & 32.22 & 29.16 & 27.40 & 31.18 & 28.29 & 26.84 \\
		& (0.9537) & (0.9093) & (0.8619) & (0.9034) & (0.8196) & (0.7518) & (0.8855) & (0.7840) & (0.7106) \\
		\hline
		\multirow{2}{*}{SCN \cite{wang2015deep}} &
		\textcolor{blue}{36.93} & \textcolor{blue}{33.10} & \textcolor{blue}{30.86} & \textcolor{blue}{32.56} & \textcolor{blue}{29.41} &
		\textcolor{blue}{27.64} & \textcolor{blue}{31.40} & \textcolor{blue}{28.50} & \textcolor{blue}{27.03} \\
		& \textcolor{blue}{(0.9552)} & \textcolor{blue}{(0.9144)} & \textcolor{blue}{(0.8732)} & \textcolor{blue}{(0.9074)} &
		\textcolor{blue}{(0.8238)} & \textcolor{blue}{(0.7578)} & \textcolor{blue}{(0.8884)} & \textcolor{blue}{(0.7885)} & \textcolor{blue}{(0.7161)} \\
		\hline
		\multirow{2}{*}{MSCN-4} &
		\textcolor{red}{37.16} & \textcolor{red}{33.33} & \textcolor{red}{31.08} & \textcolor{red}{32.85} & \textcolor{red}{29.65} &
		\textcolor{red}{27.87} & \textcolor{red}{31.65} & \textcolor{red}{28.66} & \textcolor{red}{27.19} \\
		& \textcolor{red}{(0.9565)} & \textcolor{red}{(0.9155)} & \textcolor{red}{(0.8740)} & \textcolor{red}{(0.9084)} &
		\textcolor{red}{(0.8272)} & \textcolor{red}{(0.7624)} & \textcolor{red}{(0.8928)} & \textcolor{red}{(0.7941)} & \textcolor{red}{(0.7229)} \\
		\hline
		\hline
		Our &       0.23 & 0.23 & 0.22 & 0.29 & 0.24 & 0.23 & 0.25 & 0.16 & 0.16 \\
		Improvement & (0.0013) & (0.0011) & (0.0008) & (0.0010) & (0.0034) & (0.0046) & (0.0044) & (0.0056) & (0.0068) \\
		\hline
	\end{tabular}
	}
\end{table*}

\begin{figure*}
	\center
	\begin{tabular}{p{2mm}@{\hskip 2mm}c@{\hskip 1mm}c@{\hskip 1mm}c}
		\rotatebox{90}{\hspace{10mm} A+ \cite{timofte2014a+}} &
		\includegraphics[height=0.27\linewidth, viewport= 0 200 529 656, clip]{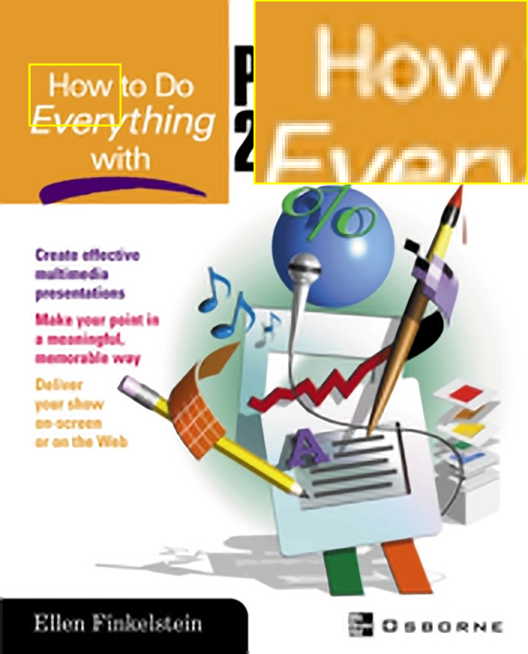} &
		\includegraphics[height=0.27\linewidth, viewport=0 200 321 481, clip]{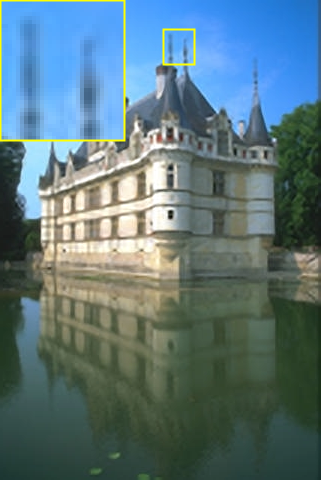} &
		\includegraphics[height=0.27\linewidth]{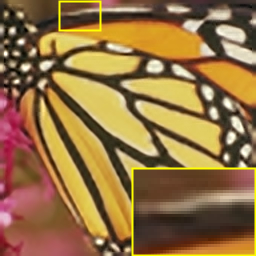} \\
		\rotatebox{90}{\hspace{7mm} SRCNN \cite{dong2015image}} &
		\includegraphics[height=0.27\linewidth, viewport= 0 200 529 656, clip]{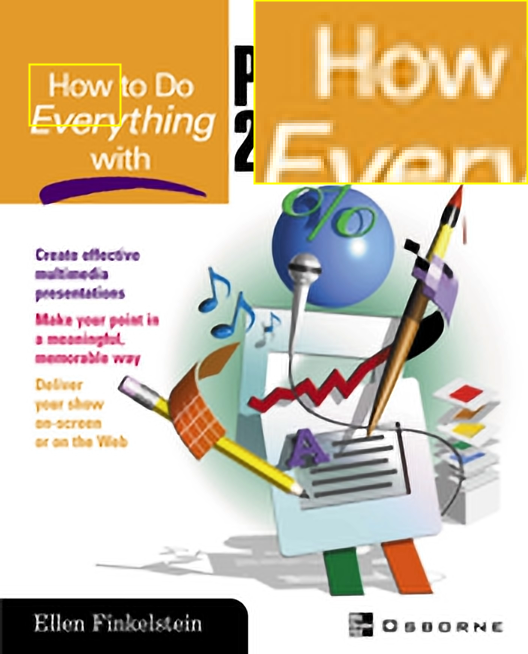} &
		\includegraphics[height=0.27\linewidth, viewport=0 200 321 481, clip]{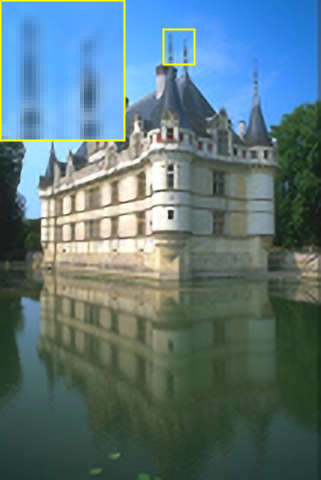} &
		\includegraphics[height=0.27\linewidth]{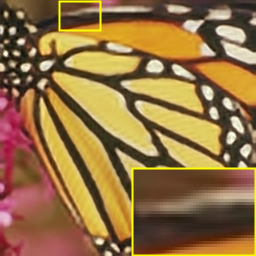} \\
		\rotatebox{90}{\hspace{8mm} SCN \cite{wang2015deep}} &
		\includegraphics[height=0.27\linewidth, viewport= 0 200 529 656, clip]{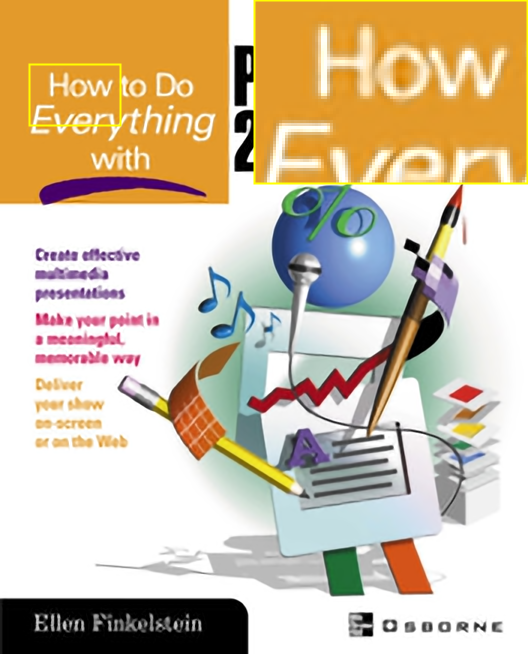} &
		\includegraphics[height=0.27\linewidth, viewport=0 200 321 481, clip]{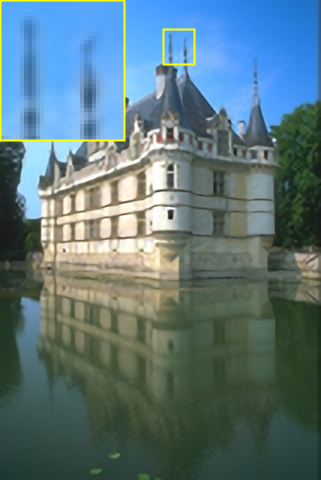} &
		\includegraphics[height=0.27\linewidth]{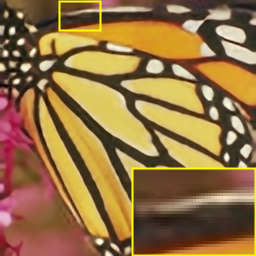} \\
		\rotatebox{90}{\hspace{8mm} MSCN-4} &
		\includegraphics[height=0.27\linewidth, viewport= 0 200 529 656, clip]{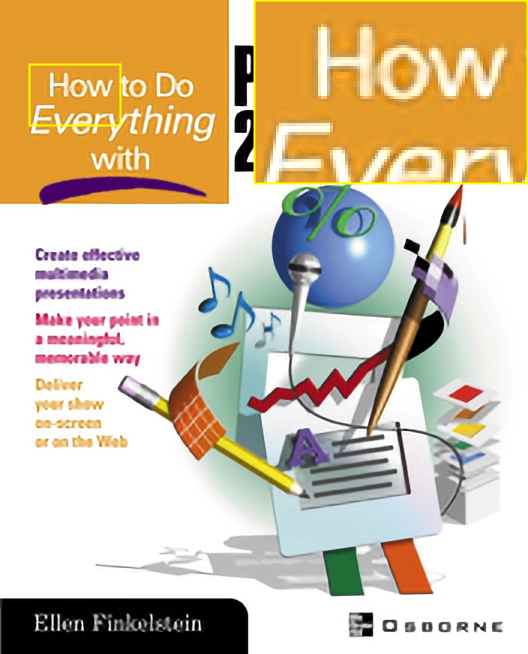} &
		\includegraphics[height=0.27\linewidth, viewport=0 200 321 481, clip]{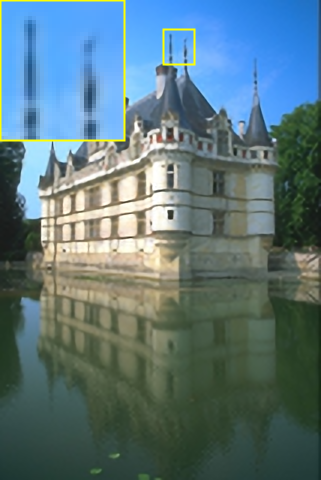} &
		\includegraphics[height=0.27\linewidth]{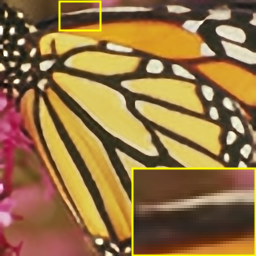} \\
		\rotatebox{90}{\hspace{3.5mm} Ground Truth} &
		\includegraphics[height=0.27\linewidth, viewport= 0 200 529 656, clip]{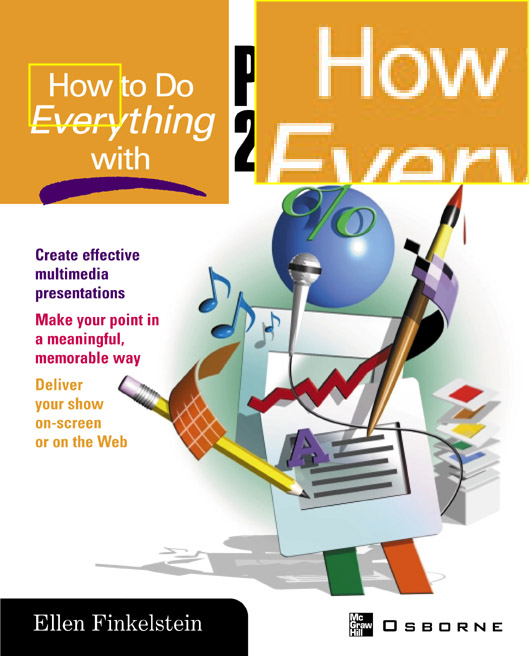} &
		\includegraphics[height=0.27\linewidth, viewport=0 200 321 481, clip]{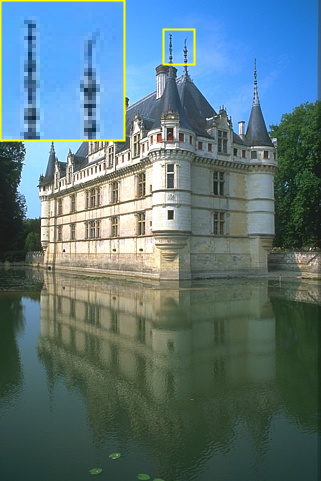} &
		\includegraphics[height=0.27\linewidth]{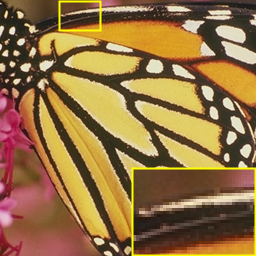} \\
		 & (a) \textit{ppt3} & (b) \textit{102061} &		
		 (c) \textit{butterfly}
	\end{tabular}
	\caption{ Visual comparisons of SR results among different methods. 
		From left to right: 
		the \textit{ppt3} image upscaled by $\times 3$; the \textit{102061} image upscaled by $\times 3$; the \textit{butterfly} image upscaled by $\times 4$.}
	\label{fig:visual}
\end{figure*}

\subsection{Comparison with State-of-the-Arts}

We conduct experiments on all the images in Set5, Set14 and BSD100 for different upscaling factors ($\times 2, \times 3, $ and $\times 4$), to quantitatively and qualitatively compare our own approach with a number of state-of-the-arts image SR methods. Table \ref{tab:psnr2} shows the PSNR and SSIM for adjusted anchored neighborhood regression (A+) \cite{timofte2014a+}, SRCNN \cite{dong2015image}, RFL \cite{schulter2015fast}, SelfEx \cite{huang2015single} and our proposed model, \textit{MSCN-4 (n=128) }, that consists of four SCN modules with $n=128$. The single generic SCN without multi-view testing in \cite{wang2015deep}, i.e.  \textit{SCN (n=128)}   is also included for comparison as the baseline. 
Note that all the methods use the same 91 images \cite{yang2010image} for training except  SRCNN \cite{dong2015image}, which uses 395,909 images from ImageNet as training data.

It can be observed that our proposed model achieves the best SR performance consistently over three data sets for various upscaling factors.
It outperforms \textit{SCN (n=128)} which obtains the second best results by about 0.2dB across all the data sets, owing to the power of multiple inference modules. 

We compare the visual quality of SR results among various methods in Figure  \ref{fig:visual}. 
The region inside the bounding box is zoomed in and shown for the sake of visual comparison. 
Our proposed model \textit{MSCN-4 (n=128) } is able to recover sharper edges and generate less artifacts in the SR inferences.

\begin{figure}[t]
	\center
	\includegraphics[width=0.9\linewidth]{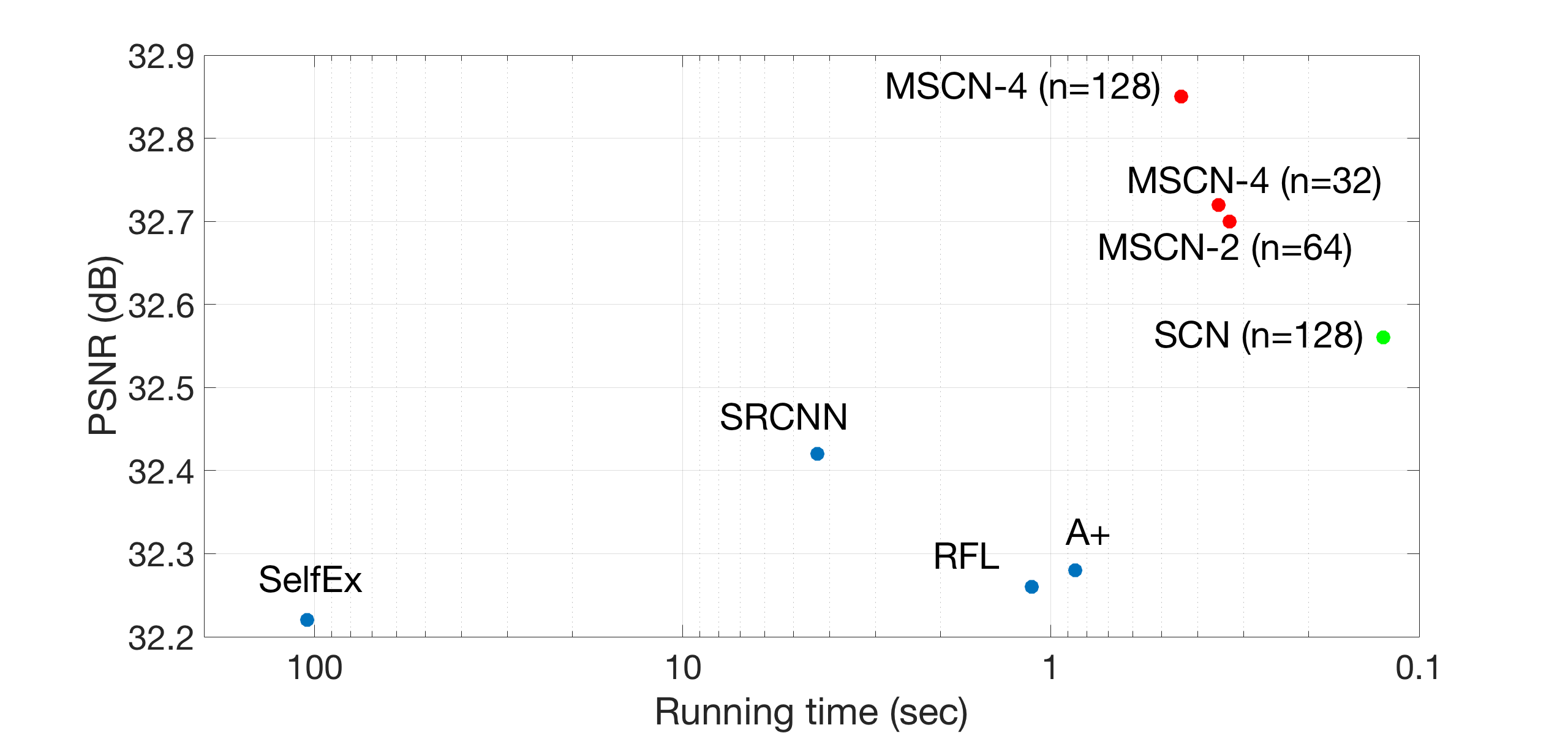}
	\caption{The average PSNR and the average inference time for upscaling factor $\times 2$ on Set14  are compared among different network structures of our method and other SR methods. SRCNN uses the public slower implementation of CPU.}
	\label{fig:runningtime}
\end{figure}

\subsection{SR Performance vs. Inference Time}

The inference time is an important factor of SR algorithms other than the SR performance. The relation between the SR performance and the inference time of our approach is analyzed in this section. 
Specifically, we measure the average inference time of different network structures in our method for upscaling factor $\times 2$ on Set14. The inference time costs versus the PSNR values are displayed in Figure \ref{fig:runningtime}, where several other current SR methods \cite{huang2015single,schulter2015fast,dong2015image,timofte2014a+} are included as reference (the inference time of SRCNN is from the public slower implementation of CPU).  
We can see that generally, the more modules our network has, the more inference time is needed and the better SR results are achieved. 
By adjusting the number of SR inference modules in our network structure, we can achieve the tradeoff between SR performance and computation complexity.
However, our slowest network still has the superiority in term of inference time, compared with other previous SR methods.

\section{Conclusions}

\label{sec:conclusions}
In this paper, we propose to jointly learn a mixture of deep networks for single image super-resolution, each of which serves as a SR inference module to handle a certain class of image signals. An adaptive weight module is designed to  predict pixel-level aggregation weights of the HR estimates. Various network architectures are analyzed in terms of the SR performance and the inference time, which validates the effectiveness of our proposed model design. Extensive experiments manifest that our proposed model is able to achieve outstanding SR performance along with more flexibility of design. In the future, this approach of image super-resolution will be explored to facilitate other high-level vision tasks \cite{wang2016studying}.


\bibliographystyle{splncs}
\bibliography{egbib}


\end{document}